\pdfoutput=1

\documentclass{article}

\usepackage[nonatbib,preprint]{neurips_2020}
\usepackage{url}
\usepackage{booktabs}
\usepackage{amsfonts}
\usepackage{nicefrac}
\usepackage{microtype}
\usepackage{xspace}
\usepackage[pdftex]{graphicx}
\usepackage{subfigure,wrapfig}
\usepackage[pdftex]{color}
\usepackage{cite}
\usepackage{amsmath}
\usepackage[ruled]{algorithm2e}

\title{Scaling Up Deep Neural Network Optimization for Edge Inference}

\author{
 Bingqian Lu\thanks{E-mail: blu029@ucr.edu}\\
  UC Riverside\\
   \And
  Jianyi Yang\thanks{E-mail: jyang239@ucr.edu}\\
  UC Riverside \\
   \And
  Shaolei Ren\thanks{E-mail: sren@ece.ucr.edu}\\
 UC Riverside\\
}

\begin{document}

\maketitle

\begin{abstract}
Deep neural networks (DNNs) have been increasingly deployed on and integrated
with edge devices, such as mobile phones, drones, robots and wearables.
To run DNN inference directly on edge devices (a.k.a. edge inference) with a satisfactory performance,
optimizing the DNN design (e.g., network architecture and quantization policy) is crucial. While
state-of-the-art DNN designs have leveraged performance predictors
to speed up the optimization process, they are device-specific (i.e., each predictor for only one
target device) and hence cannot
 scale well in the presence of extremely diverse edge devices.
Moreover, even with performance predictors, the optimizer (e.g., search-based
optimization) can still be time-consuming when optimizing DNNs for many
different devices.
 In this work, we propose two approaches
 to scaling up DNN optimization.
 In the first approach,
 we reuse the performance predictors
built on a \emph{proxy} device, and leverage
the performance monotonicity to scale up the DNN optimization without re-building performance
predictors for each different device.
In the second approach,
we build scalable performance predictors that can
estimate the resulting  performance (e.g., inference accuracy/latency/energy)
given a DNN-device pair,
and use a neural network-based automated optimizer that takes both device features
and optimization parameters as input and then directly outputs the optimal
DNN design without going through a lengthy optimization process for each individual
device.
\end{abstract}

\section{Background and Motivation}

Deep neural networks (DNNs) have been increasingly deployed on and integrated
with edge devices, such as mobile phones, drones, robots and wearables.
Compared
to cloud-based inference, running DNN inference directly
on edge devices (a.k.a. \emph{edge inference})
has several major advantages, including
being free from the network connection requirement, saving
bandwidths and better protecting
user privacy as a result of local data processing.
For example, it is very common to include one or multiple
DNNs in today's mobile apps \cite{DNN_Facebook_Inference_HPCA_2019}.

To achieve a satisfactory user experience for edge inference,
an appropriate DNN design is needed to optimize a multi-objective performance
metric, e.g.,
 good accuracy while keeping the latency and energy consumption low.
A complex DNN model involves multi-layer perception with up to billions
of parameters, imposing a stringent computational and memory
requirement that is often too prohibitive for edge devices.
Thus,
 the DNN models
running on an edge device
must
be judiciously optimized using, e.g., neural architecture search
(NAS) and model compression \cite{DNN_Compression_AutoCompress_YanzhiWang_AAAI_2020,DNN_NAS_APQ_JointSearch_ArchitecturePruningQuantization_SongHan_CVPR_2020_Wang_2020_CVPR,DNN_NAS_OnceForAll_SongHan_MIT_2020_ICLR,DNN_NAS_Hardware_YiyuShi_ICCAD_2019_Jiang:2019:AVE:3316781.3317757,DNN_NAS_NeuralPower_PredictEnergy_CMU_ACML_2017_pmlr-v77-cai17a,DNN_NAS_PlatformAware_Mobile_MeasureLatency_Google_CVPR_2019_tan2019mnasnet,DNN_NAS_ProxylessNAS_LatencyPredictor_SongHan_ICLR_2019_cai2018proxylessnas}.

The DNN design choices we focus on in this work mainly refer to the network architecture
 and compression scheme (e.g., pruning and quantization policy), which constitute an
 exponentially large space. Note that the other DNN design parameters, such as learning rate
 and choice of optimizer  for
 DNN training, can also be included into the proposed framework.
 For example, if we want to consider
 learning rate and DNN architecture optimization, the accuracy predictor can take the
 learning rate and architecture as the input and
 be trained by using different DNN samples with distinct architectures and learning rates.

 Given different design choices, DNN models can exhibit dramatically
 different  performance tradeoffs in terms
 of various important performance metrics (e.g., accuracy, latency, energy and robustness).
In general, there is not a single DNN model that performs Pareto optimally on all edge devices.
For example, with the same DNN model in Facebook's app, the resulting
latencies on different devices can vary significantly \cite{DNN_Facebook_Inference_HPCA_2019}.
Thus,  device-aware DNN optimization is mandated \cite{DNN_NAS_Hardware-awareTransformer_Lanugage_SongHan_ACL_2020_wang2020hat,DNN_NAS_Hardware_YiyuShi_ICCAD_2019_Jiang:2019:AVE:3316781.3317757,DNN_Facebook_Inference_HPCA_2019,DNN_Systems_ShuffleNetV2_Guideline_ArchitectureDesign_10.1007/978-3-030-01264-9_8}.

\begin{wrapfigure}[]{r}{0.4\textwidth}
	\includegraphics[trim=0cm 0.0cm 0cm 0.0cm, clip,width=0.4\textwidth]{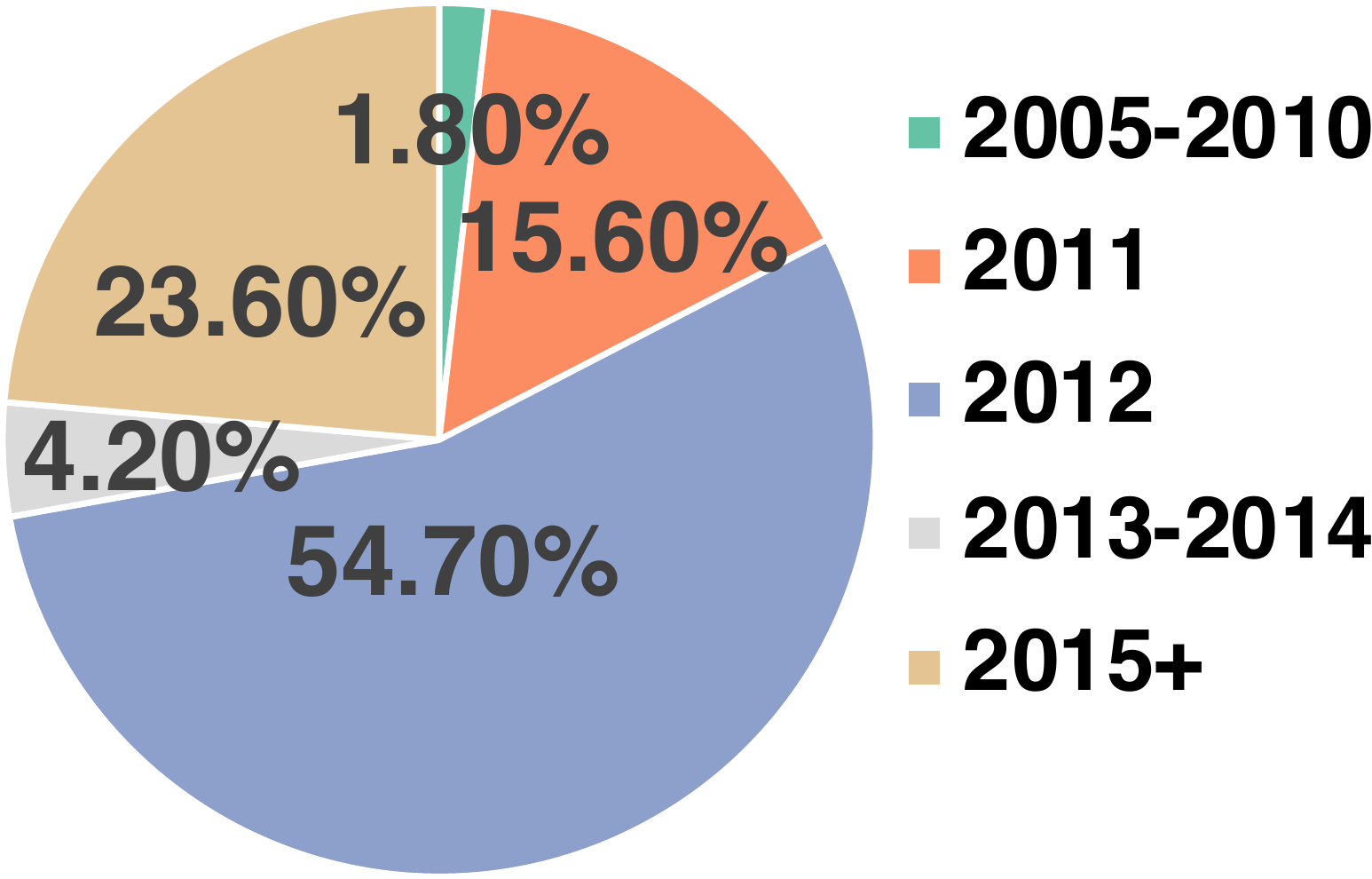}
	\caption{Statistics of the year mobile CPUs are designed as of late 2018 \cite{DNN_Facebook_Inference_HPCA_2019}.}
	\label{fig:CPU_soc}
\end{wrapfigure}
Designing an optimal DNN for even a single edge
device often needs repeated design iterations and is non-trivial \cite{DNN_NAS_MSNet_StructuralSearch_IoT_YiranChen_ICCV_2019_Cheng_2019_ICCV,DNN_FBNet_HardwareAwareConvNetDesign_CVPR_2019_Wu2018FBNetHE}.
Worse yet, DNN model developers often need to serve extremely diverse
edge devices. For example,  the DNN-powered voice assistant application
developed by a third party can be used by many different
edge device vendors,
and Facebook's DNN model
for style transfer is run on billions of mobile devices,
more than half of which still
use CPUs designed in 2012 or before (shown in Fig.~\ref{fig:CPU_soc}) \cite{DNN_Facebook_Inference_HPCA_2019}.
In the mobile market alone, there are thousands
of system-on-chips (SoCs) available. Only top 30 SoCs can each
take up more than 1\% of the share,
and they collectively account for
 51\% of the whole market \cite{DNN_Facebook_Inference_HPCA_2019}.
Thus, the practice of repeatedly optimizing DNN models, once
for each edge device, can no longer meet the demand in view of the extremely diverse edge
devices.

Therefore,
it has become crucially important to scale up
the optimization of DNNs for edge inference using automated approaches.

\section{State of the Art and Limitations}

Network architecture is a key design choice
that affects the resulting performance of DNN models on edge devices.
Due to the huge space for network architectures,
traditional
hand-tuned architecture designs can take months or even longer
to train a DNN with a satisfactory performance \cite{DNN_NAS_RL_ArchitectureSearch_ICLR_2017_zoph2016neural,DNN_NAS_Survey_2019_elsken2019neural}. Thus,
they have become obsolete and been replaced with automated
approaches \cite{DNN_NAS_PlatformAware_Mobile_MeasureLatency_Google_CVPR_2019_tan2019mnasnet}.
Nonetheless, the early NAS approaches
often require training each DNN candidate
 (albeit usually on a small proxy dataset),
which hence still results in a high complexity and search time.
To address this issue, DNN optimization and training need
to be decoupled. For example, the current ``once-for-all'' technique can generate
nearly unlimited ($>10^{19}$) DNN models of different architectures all at once \cite{DNN_NAS_OnceForAll_SongHan_MIT_2020_ICLR}. Consequently,
DNN model developers can now focus on the optimization of network architecture, without having
to train a DNN for each candidate architecture.
Thus, instead of DNN training, we consider on scalability of optimizing DNN
designs with a focus on the neural architecture.

NAS on a single target device cannot result in the optimal DNN model
for all other devices, motivating device-aware NAS.
In general, the device-aware
NAS process is guided by an objective
function, e.g., $accuracy\_loss + weight_1*energy + weight_2*latency$.
Thus, it is crucial to efficiently evaluate the
resulting inference accuracy/latency/energy performance
given a DNN candidate \cite{DNN_NAS_Predictor_AlphaX_2019_wang2019alphax,DNN_NAS_Predictor_BayesianOptimization_Oxford_2020_ru2020neural,DNN_NAS_Predictor_Survey_YuWang_Tsinghua_2020_ning2020surgery,DNN_NAS_Predictor_NAO_NIPS_2018_luo2018neural,DNN_NAS_Predictor_TongZhang_2019_shi2019multi}.
Towards this end,
proxy models have been leveraged to calculate latency/energy for each candidate,
but they are
 not very accurate on all devices \cite{DNN_FBNet_HardwareAwareConvNetDesign_CVPR_2019_Wu2018FBNetHE}.
Alternatively, actual latency measurement
on real devices for each candidate is also considered, but it  is time-consuming \cite{DNN_NAS_PlatformAware_Mobile_MeasureLatency_Google_CVPR_2019_tan2019mnasnet}.

More recently, performance predictors or lookup tables  have been utilized
to assist with NAS (and model compression) \cite{DNN_NAS_Predictor_AlphaX_2019_wang2019alphax,DNN_NAS_Predictor_BayesianOptimization_Oxford_2020_ru2020neural,DNN_NAS_Predictor_Survey_YuWang_Tsinghua_2020_ning2020surgery,DNN_NAS_Predictor_NAO_NIPS_2018_luo2018neural,DNN_NAS_Predictor_TongZhang_2019_shi2019multi,DNN_NAS_Hardware_YiyuShi_ICCAD_2019_Jiang:2019:AVE:3316781.3317757,DNN_NAS_HyperPower_Predictor_CMU_DATE_2018_8341973,DNN_NAS_APQ_JointSearch_ArchitecturePruningQuantization_SongHan_CVPR_2020_Wang_2020_CVPR,DNN_NAS_NeuralPower_PredictEnergy_CMU_ACML_2017_pmlr-v77-cai17a}:
train a machine learning model or build a lookup table to estimate the resulting
accuracy/latency/energy performance for a candidate DNN design on the target device.
Therefore, by using search techniques aided by
performance predictors or lookup tables, an optimal DNN can be
identified out of numerous candidates for a target edge device
without
actually deploying or running  each candidate DNN   on the device \cite{DNN_NAS_APQ_JointSearch_ArchitecturePruningQuantization_SongHan_CVPR_2020_Wang_2020_CVPR,DNN_NAS_OnceForAll_SongHan_MIT_2020_ICLR}.

\begin{figure}[!t]
    \centering
	\includegraphics[trim=0cm 10cm 8.5cm 0.0cm, page=1,clip,width=0.8\textwidth]{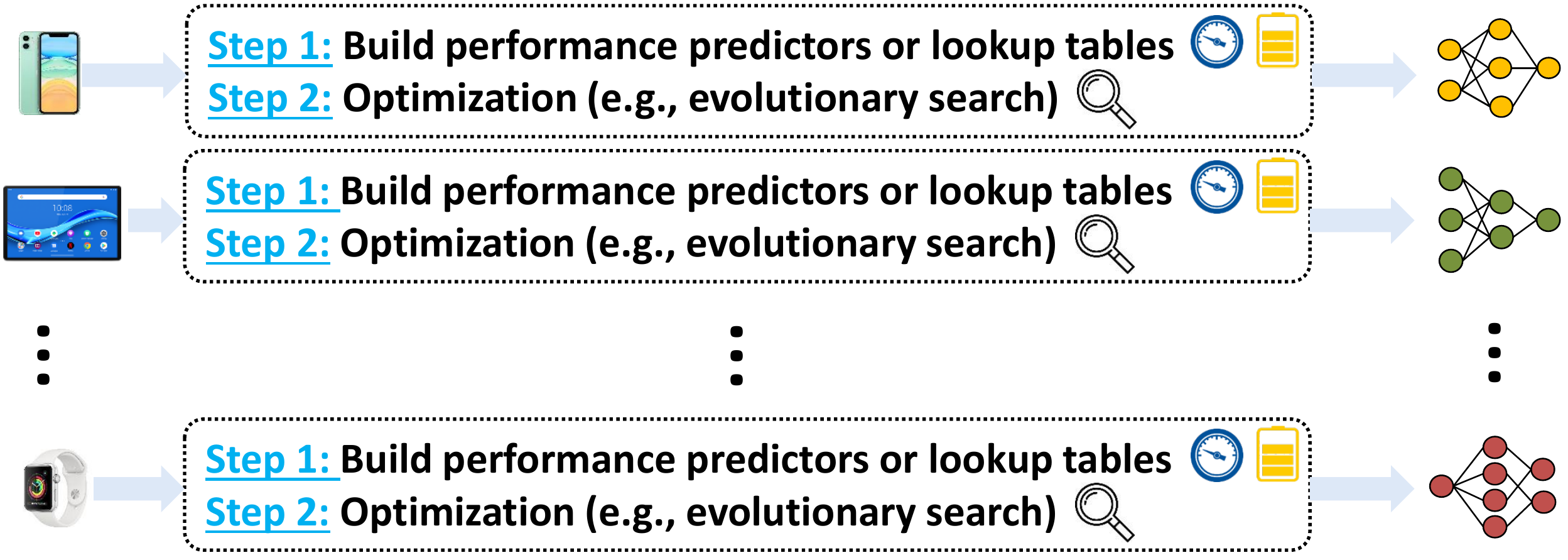}
	\caption{The existing 
device-aware DNN optimization (i.e., once  for a single device) \cite{DNN_NAS_OnceForAll_SongHan_MIT_2020_ICLR,DNN_NAS_APQ_JointSearch_ArchitecturePruningQuantization_SongHan_CVPR_2020_Wang_2020_CVPR,DNN_NAS_ChamNet_Prediction_CVPR_2019_dai2019chamnet}.}
	\label{fig:existing}
\end{figure}

Nonetheless,
as illustrated in Fig.~\ref{fig:existing},
 the existing latency/energy predictors or lookup tables  \cite{DNN_NAS_Predictor_Survey_YuWang_Tsinghua_2020_ning2020surgery,DNN_NAS_Predictor_TongZhang_2019_shi2019multi,DNN_NAS_ChamNet_Prediction_CVPR_2019_dai2019chamnet,DNN_NAS_ProxylessNAS_LatencyPredictor_SongHan_ICLR_2019_cai2018proxylessnas,DNN_NAS_APQ_JointSearch_ArchitecturePruningQuantization_SongHan_CVPR_2020_Wang_2020_CVPR,DNN_NAS_OnceForAll_SongHan_MIT_2020_ICLR}
are \emph{device-specific} and only take the DNN features as input
to predict the inference latency/energy performance on a particular target device.
For example,
according to  \cite{DNN_NAS_ProxylessNAS_LatencyPredictor_SongHan_ICLR_2019_cai2018proxylessnas}, the average inference latencies of 4k randomly selected sample DNNs
 are measured on a  mobile device
and then used
to train an average latency predictor for that specific device (plus additional 1k samples for testing).
Assuming that each measurement takes 30 seconds, {it takes a total of 40+ hours to just collect training and testing samples in order to building the latency predictor for one single device}, let alone
the additional time spent for latency predictor training and other performance predictors.
Likewise, to estimate
the inference latency,
350K operator-level latency records are profiled to construct a lookup table in \cite{DNN_NAS_ChamNet_Prediction_CVPR_2019_dai2019chamnet},
which is inevitably time-consuming.
Clearly, building performance predictors or lookup tables
 incurs a significant overhead by itself \cite{DNN_NAS_Predictor_Survey_YuWang_Tsinghua_2020_ning2020surgery,DNN_NAS_Predictor_TongZhang_2019_shi2019multi,DNN_NAS_ChamNet_Prediction_CVPR_2019_dai2019chamnet,DNN_NAS_ProxylessNAS_LatencyPredictor_SongHan_ICLR_2019_cai2018proxylessnas,DNN_NAS_APQ_JointSearch_ArchitecturePruningQuantization_SongHan_CVPR_2020_Wang_2020_CVPR,DNN_NAS_OnceForAll_SongHan_MIT_2020_ICLR}.

More crucially, without taking into account the device features, the resulting performance predictors or lookup tables only
provide good predictions for
the individual device on which the performance is measured.
For example,
 as shown in Fig.~4 in \cite{DNN_NAS_ChamNet_Prediction_CVPR_2019_dai2019chamnet},
the same convolution operator can result
in dramatically different latencies on two different devices --- Samsung S8
with Snapdragon 835 mobile CPU and
Hexagon v62 DSP with 800 MHz frequency.

In addition, the optimizer (e.g., a simple evolutionary search-based algorithm
or more advanced exploration strategies \cite{DNN_NAS_Predictor_TongZhang_2019_shi2019multi,DNN_NAS_Predictor_BayesianOptimization_Oxford_2020_ru2020neural,DNN_NAS_Predictor_NAO_NIPS_2018_luo2018neural,DNN_NAS_Predictor_Survey_YuWang_Tsinghua_2020_ning2020surgery})
to identify an optimal architecture  for each device also takes non-negligible time or CPU-hours.
For example, even with limited rounds of evolutionary search, 30 minutes to several hours
are needed by the DNN optimization process for each device \cite{DNN_NAS_OnceForAll_SongHan_MIT_2020_ICLR,DNN_NAS_APQ_JointSearch_ArchitecturePruningQuantization_SongHan_CVPR_2020_Wang_2020_CVPR,DNN_NAS_StandingShoulder_YiyuShi_HardwareCoDesign_TCAD_2020_jiang2020standing}.
In \cite{DNN_NAS_ChamNet_Prediction_CVPR_2019_dai2019chamnet}, the
search time may reduce to a few minutes by only searching for similar architectures
compared to an already well-designed baseline DNN model, and hence this comes at the expense
of very limited search space and  possibly missing better DNN designs.
Therefore,
combined together, the total search cost for edge devices is still non-negligible, especially given the extremely diverse edge devices for which scalability is very important.

There have also been many prior studies on DNN model compression,
such as pruning and quantization \cite{DNN_Compression_CirCNN_XuehaiQian_Micro_2017_10.1145/3123939.3124552,DNN_Compression_AutoCompress_YanzhiWang_AAAI_2020,DNN_Compression_PatDNN_Mobile_YanzhiWang_ASPLOS_2020,DNN_Compression_SongHan_ICLR_2016,model_compression_survey, songhan_learning_weights_connections, fused_layer_CNN_accelerator, train_DNN_with_binary_weights, imagenet_classification_binary_CNN, embedded_binarized_NN}, matrix factorization \cite{CNN_linear_structure, surveillance_as_edge_network_service}, and knowledge distillation \cite{knowledge_transfer_for_DL_at_edge}, among  others.
Like the current practice of NAS, the existing optimizer
for compression techniques
are typically targeting a single device
(e.g., optimally deciding the quantization and pruning policy
for an individual target device), thus
making the overall optimization cost linearly increase with the
number of target devices and lacking scalability \cite{DNN_NAS_APQ_JointSearch_ArchitecturePruningQuantization_SongHan_CVPR_2020_Wang_2020_CVPR}.

In summary,
 the state-of-the-art device-aware DNN optimization  still takes
a large amount of time and efforts
for even a single device \cite{DNN_NAS_ProxylessNAS_LatencyPredictor_SongHan_ICLR_2019_cai2018proxylessnas,DNN_NAS_OnceForAll_SongHan_MIT_2020_ICLR,DNN_NAS_APQ_JointSearch_ArchitecturePruningQuantization_SongHan_CVPR_2020_Wang_2020_CVPR,DNN_NAS_ChamNet_Prediction_CVPR_2019_dai2019chamnet},
and  cannot scale to extremely diverse edge devices.

\section{Problem Formulation}

A common goal of optimizing DNN designs is to maximize
the inference accuracy subject to latency and/or energy constraints on edge devices.
Mathematically,
this problem
can be formulated as
\begin{eqnarray}\label{eqn:objective_original}
  &&\min_{\mathbf{x}\in\mathcal{X}} -accuracy(\mathbf{x})\\
  \label{eqn:constraint_latency}
  s.t., &&latency(\mathbf{x};\mathbf{d})\leq \overline{L}_{\mathbf{d}},\\
  \label{eqn:constraint_energy}
  && energy(\mathbf{x};\mathbf{d})\leq \overline{E}_{\mathbf{d}},
\end{eqnarray}
where $\mathbf{x}$ is the representation of the DNN design choice (e.g.,
a combination of DNN architecture, quantization, and pruning scheme), $\mathcal{X}$
is the design space under consideration,
and $\mathbf{d}$ is the representation of an edge device (e.g.,  CPU/RAM/GPU/OS configuration). Our problem formulation is not restricted
to energy and latency constraints; additional constraints,
such as robustness to adversarial samples, can also be added.
Note that we use ``$-accuracy(\mathbf{x})$'' as the objective
function to be consistent with the standard ``$\min$'' operator in optimization problems.

The constrained optimization problem in Eqns.~\eqref{eqn:objective_original}--\eqref{eqn:constraint_energy} is called
\emph{primal} problem in the optimization literature \cite{BoydVandenberghe}.
It can also be alternatively formulated as a relaxed problem parameterized by
$\lambda=(\lambda_1, \lambda_2)$:
\begin{equation}\label{eqn:obj_transformed}
\min_{\mathbf{x}\in\mathcal{X}}-accuracy(\mathbf{x})+\lambda_1\cdot energy(\mathbf{x};\mathbf{d}) + \lambda_2\cdot latency(\mathbf{x};\mathbf{d}),
\end{equation}
where $\lambda=(\lambda_1, \lambda_2)$ are non-negative weight parameters (i.e.,
equivalent to Lagrangian multipliers) corresponding
to the energy and latency constraints, respectively. By increasing
a weight  (say, $\lambda_2$ for latency),
the optimal design $\mathbf{x}^*(\mathbf{d},\lambda)$ by solving \eqref{eqn:obj_transformed}
will result in better performance corresponding to that weight.
If the performance constraint
is very loose, then $\lambda=(\lambda_1, \lambda_2)$ can approach zero;
on the other hand, if the constraint is very stringent, $\lambda=(\lambda_1, \lambda_2)$
will be large.
Thus, given
a set of latency and energy constraints $\overline{L}_{\mathbf{d}}$
and $\overline{E}_{\mathbf{d}}$, we can choose a set of weight parameters
$\lambda_1$ and $\lambda_2$ such that the constraints in \eqref{eqn:constraint_latency}\eqref{eqn:constraint_energy} are satisfied
and the accuracy is maximized.


Strictly speaking,
some technical conditions (e.g., convexity)
need to be satisfied such that the
optimal solution to the relaxed problem
in \eqref{eqn:obj_transformed} is also the optimal solution to the constrained problem in
\eqref{eqn:objective_original}--\eqref{eqn:constraint_energy}.
Nonetheless,
the goal in practice is to obtain a sufficiently good DNN design
rather than the truly global optimum, because of the usage
of a (non-convex) performance predictor as a substitute of
the objective
function  \cite{DNN_NAS_OnceForAll_SongHan_MIT_2020_ICLR,DNN_NAS_ProxylessNAS_LatencyPredictor_SongHan_ICLR_2019_cai2018proxylessnas,DNN_NAS_Hardware_YiyuShi_ICCAD_2019_Jiang:2019:AVE:3316781.3317757,DNN_NAS_APQ_JointSearch_ArchitecturePruningQuantization_SongHan_CVPR_2020_Wang_2020_CVPR,DNN_NAS_ChamNet_Prediction_CVPR_2019_dai2019chamnet}.
Thus, with proper weight parameters $\lambda$,  the relaxed version in \eqref{eqn:obj_transformed} can be seen as a substitute of
the constrained optimization problem \eqref{eqn:objective_original}--\eqref{eqn:constraint_energy}.

While the constrained problem formulation in \eqref{eqn:objective_original}--\eqref{eqn:constraint_energy}
is intuitive to understand, it may not be straightforward to optimize when using search-based algorithms. On the other hand, when using the relaxed
formulation in \eqref{eqn:obj_transformed}, one needs to find an appropriate set
of weight parameters $\lambda=(\lambda_1, \lambda_2)$ to meet the performance
constraints in \eqref{eqn:constraint_latency}\eqref{eqn:constraint_energy}.
In the literature, both constrained and relaxed problems are widely considered to guide optimal DNN designs \cite{DNN_NAS_APQ_JointSearch_ArchitecturePruningQuantization_SongHan_CVPR_2020_Wang_2020_CVPR,DNN_NAS_ChamNet_Prediction_CVPR_2019_dai2019chamnet}.

In this paper, we choose to solve the relaxed problem in \eqref{eqn:obj_transformed}
while using efficient searches to identify an optimal $\lambda=(\lambda_1, \lambda_2)$
such that the performance constraints in \eqref{eqn:constraint_latency}\eqref{eqn:constraint_energy}
are satisfied and the resulting optimal DNN design $\mathbf{x}$ minimizes
the accuracy loss (i.e., maximize the accuracy).

\section{Approach 1: Reusing Performance Predictors for Many Devices}

A key bottleneck that slows down the DNN optimization process is the high
cost of building performance predictors for each device.
In our first approach, we propose to reuse the performance predictors
built on a \emph{proxy} device denoted as $\mathbf{d}_0$. While the predictor
cannot accurately estimate the performance on a different device, it maintains
performance \emph{monotonicity} (e.g., if DNN design $\mathbf{x}_A$
has a lower latency than $\mathbf{x}_B$ on the proxy device,
$\mathbf{x}_A$ should still be faster than $\mathbf{x}_B$ on a new device) in many cases. We leverage the performance monotonicity to scale up the DNN optimization without re-building performance
predictors for each different device.

\subsection{Stage 1: Training Performance Predictors on a Proxy Device}

To speed up the DNN optimization process, we need
to quickly evaluate objective function given different DNN designs.
Instead
of actually measuring the performance for each DNN design candidate
(which is time-consuming), we utilize performance predictors.
In our example,
we have accuracy/latency/energy predictors. Concretely,
the accuracy predictor can be a simple Gaussian process model as used
in \cite{DNN_NAS_ChamNet_Prediction_CVPR_2019_dai2019chamnet}
or a neural network, whose
input is the DNN design choice represented by $\mathbf{x}$, and it does not depend
on the edge device feature $\mathbf{d}$.
We denote
the trained accuracy predictor
by $Acc_{\Theta_A}(\mathbf{x})$,
where $\Theta_A$ is learnt parameter for the predictor.

On the other hand, the latency/energy
predictors depend on devices. Here, we train the latency/energy
predictors on a proxy device following
the existing studies
\cite{DNN_NAS_APQ_JointSearch_ArchitecturePruningQuantization_SongHan_CVPR_2020_Wang_2020_CVPR,DNN_NAS_ChamNet_Prediction_CVPR_2019_dai2019chamnet}.
For example, to build the latency predictor offline, we can measure the latency for each operator
in a DNN candidate and then sum up all the involved operators to obtain the total
latency. We denote the latency and energy predictors
as $\overline{latency}_{\mathbf{d}_0}(\mathbf{x})$ and $\overline{energy}_{\mathbf{d}_0}(\mathbf{x})$,
where the subscript $\mathbf{d}_0$ is to stress that the performance predictors
are only accurate (in terms of the absolute performance prediction) for the proxy device $\mathbf{d}_0$.

Given the latency/energy predictor for an edge device, one can easily follow
\cite{DNN_NAS_APQ_JointSearch_ArchitecturePruningQuantization_SongHan_CVPR_2020_Wang_2020_CVPR,DNN_NAS_ChamNet_Prediction_CVPR_2019_dai2019chamnet}
and adopt an evolutionary search process to obtain the optimal DNN design.
Nonetheless, in \cite{DNN_NAS_ChamNet_Prediction_CVPR_2019_dai2019chamnet}, the performance predictor cannot transfer to a different device, because the latency/energy performance
on one device can change dramatically on a different device:
\cite{DNN_NAS_ChamNet_Prediction_CVPR_2019_dai2019chamnet} directly uses the absolute
performance constraints $\overline{L}_{\mathbf{d}}$ and $\overline{E}_{\mathbf{d}}$
in its (modified) objective function and hence needs accurate performance prediction
for each individual device. In \cite{DNN_NAS_APQ_JointSearch_ArchitecturePruningQuantization_SongHan_CVPR_2020_Wang_2020_CVPR,DNN_NAS_OnceForAll_SongHan_MIT_2020_ICLR},
the weight parameters $\lambda=(\lambda_1, \lambda_2)$  are simply treated as hyperparameters. How to tune $\lambda=(\lambda_1, \lambda_2)$ to meet the performance constraints for a target device is not specified. Since it aims
at making weighted objective
function in \eqref{eqn:obj_transformed} as close to the true value
as possible on a target device, it needs accurate performance prediction
 for that target
device. Thus, performance predictors are needed for each individual device in
\cite{DNN_NAS_APQ_JointSearch_ArchitecturePruningQuantization_SongHan_CVPR_2020_Wang_2020_CVPR,DNN_NAS_OnceForAll_SongHan_MIT_2020_ICLR}.

Instead of building a latency/energy predictor
for each device, we will
 reuse the predictor for other devices as described in the next
subsection.

\subsection{Stage 2: Optimizing DNN Designs on New Devices}

\begin{figure*}[!t]
    \centering
	\includegraphics[trim=0cm 12cm 7.5cm 0.0cm, page=3,clip,width=0.9\textwidth]{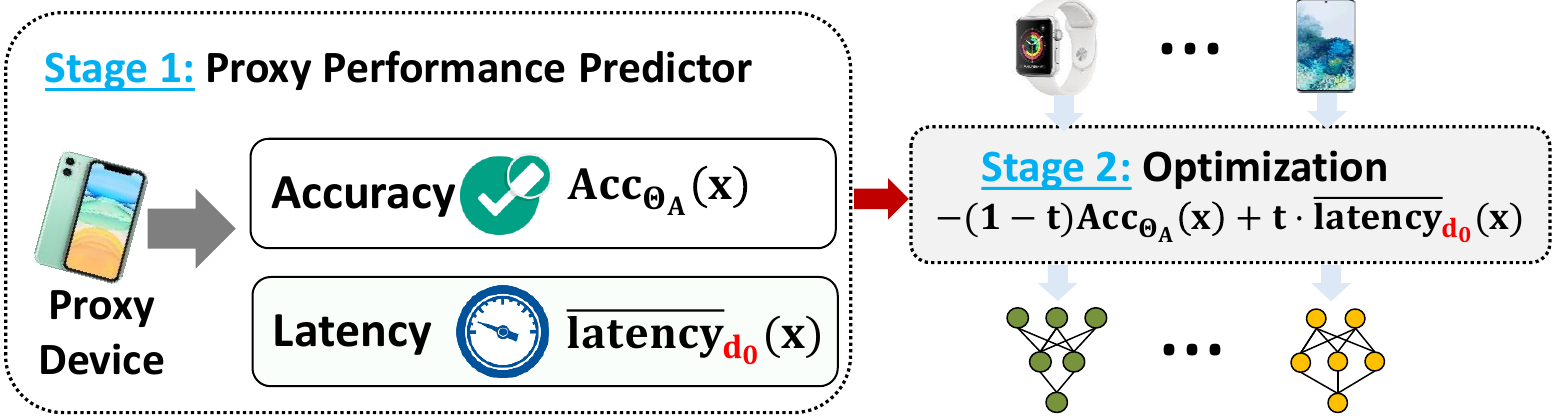}
	\caption{Overview of ``reusing performance predictors'' to scale up DNN optimization.}
	\label{fig:first_approach}
\end{figure*}
In this work, we avoid the cost of building performance predictors for each individual
device by leveraging the
performance {monotonicity} of DNNs on different devices.
To better explain our idea, we only consider the latency constraint and illustrate
our approach in Fig.~\ref{fig:first_approach}.

\begin{figure*}[!t]
    \centering
\subfigure[Latencies of 40 DNN models]{
			\includegraphics[width=0.32\textwidth]{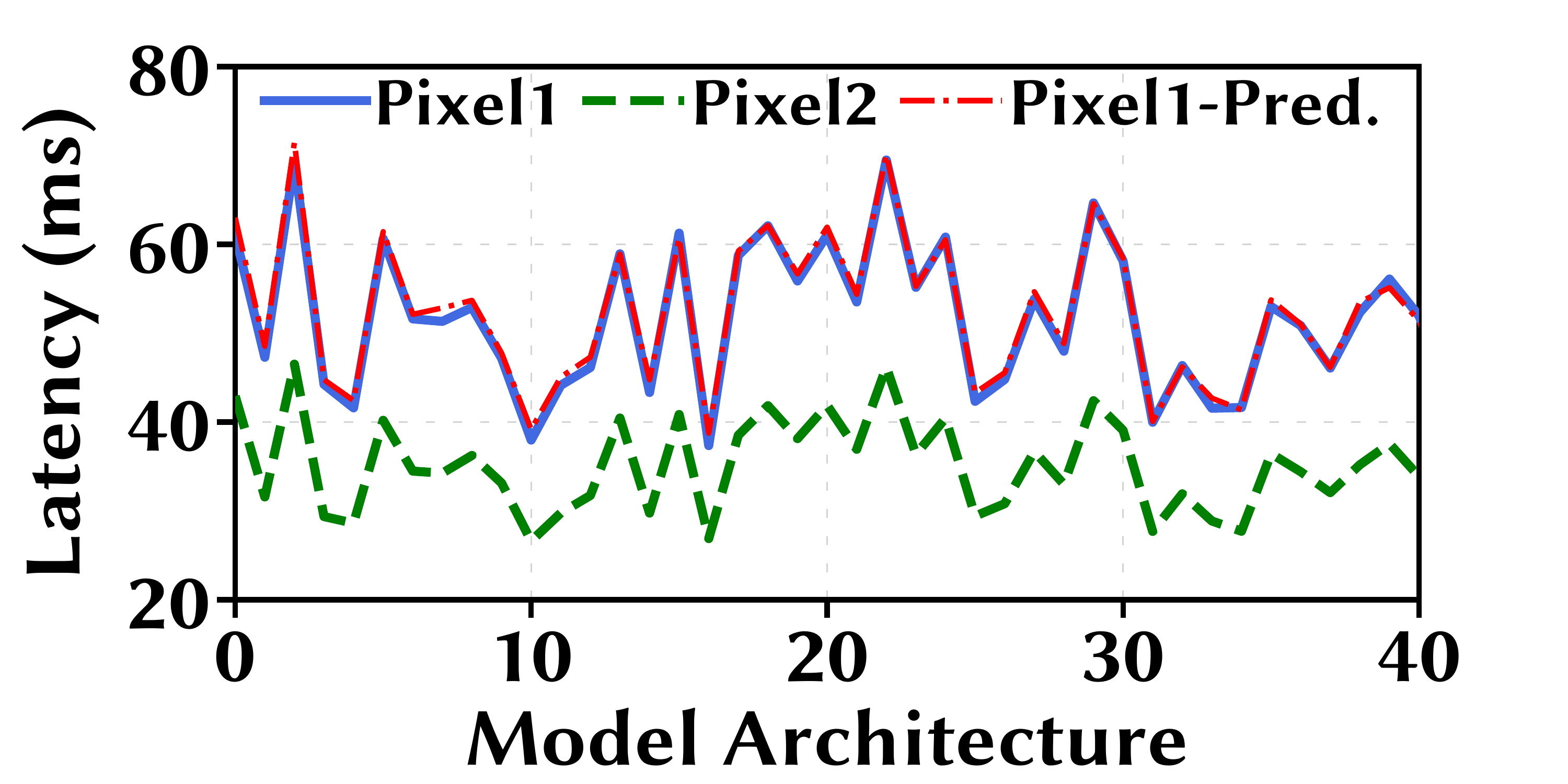}
	}%
\subfigure[Latency on Pixel 2 vs. Pixel 1]{
			\includegraphics[width=0.32\textwidth]{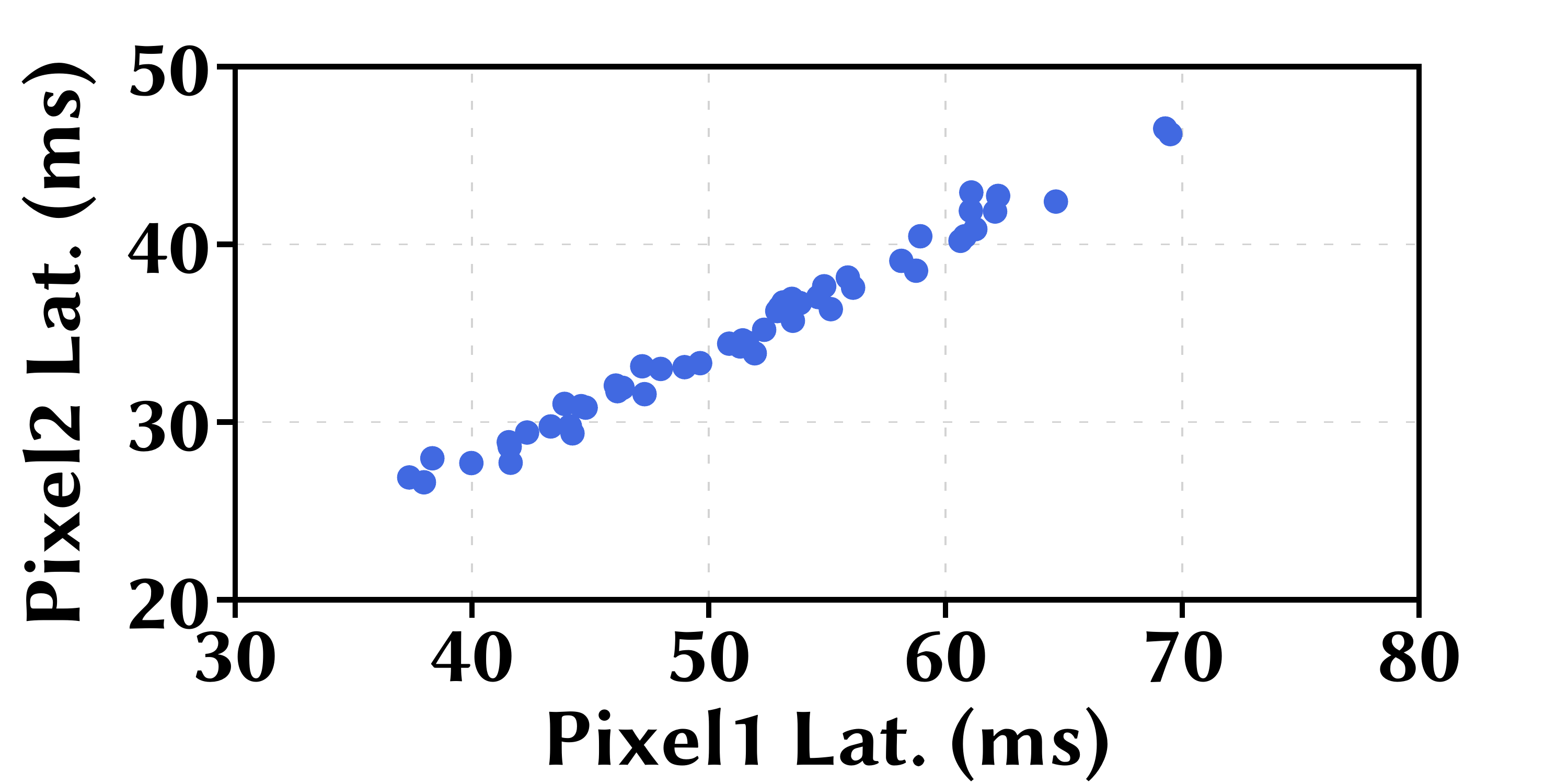}
	}%
\subfigure[Predicted vs. real latency (Pixel~1)]{
			\includegraphics[width=0.32\textwidth]{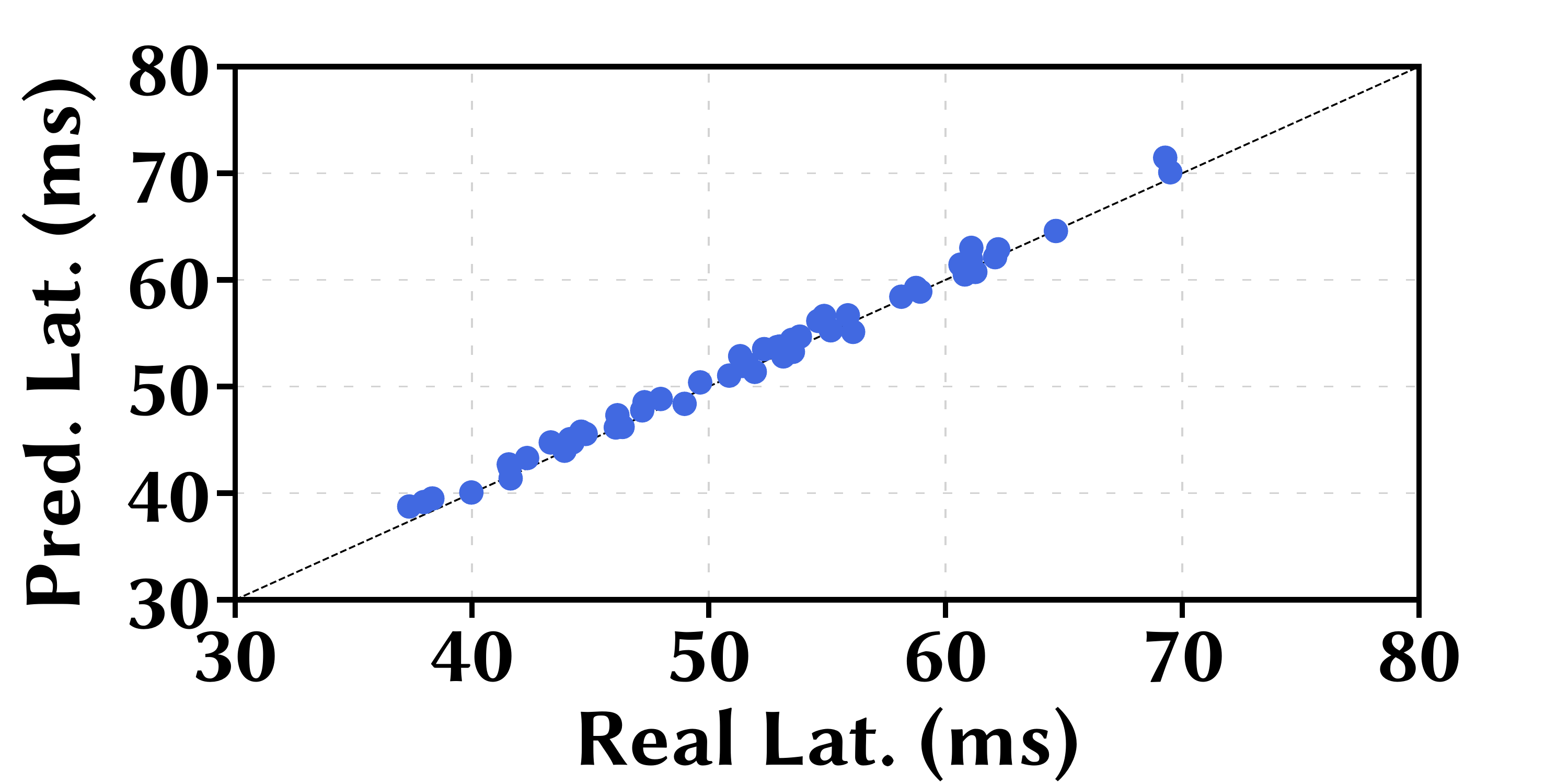}
	}	
	\caption{The measured and predicted average latencies of a set of 40 DNN models with
different architectures  on Google Pixel 1 and Pixel 2.
The latency predictor is built based on Google Pixel 1. The latency
values are released accompanying the publication \cite{DNN_NAS_ProxylessNAS_LatencyPredictor_SongHan_ICLR_2019_cai2018proxylessnas}.}
	\label{fig:performance_monotonicity}
\end{figure*}

In many cases, DNNs' latency
performances are
monotone on two different devices, which we formally state as follows.

\textbf{Performance monotonicity.}
Given two different devices $\mathbf{d}_0\not=\mathbf{d}$
and two different DNN designs $\mathbf{x}_A\not=\mathbf{x}_B$, if $latency(\mathbf{x}_A;\mathbf{d}_0)\geq
latency(\mathbf{x}_B;\mathbf{d}_0)$, then $latency(\mathbf{x}_A;\mathbf{d})\geq
latency(\mathbf{x}_B;\mathbf{d})$ also holds. We say
that the two DNN designs $\mathbf{x}_A$ and $\mathbf{x}_B$
are performance monotonic on the  two devices $\mathbf{d}_0$ and $\mathbf{d}$.

\begin{algorithm}[!t]
	\caption{DNN Optimization on a New Device}
	\textbf{Input: }
	Accuracy predictor $Acc_{\Theta_A}(\mathbf{x})$, proxy device's latency predictor $\overline{latency}_{\mathbf{d}_0}(\mathbf{x})$, latency constraint on the target device
$\overline{L}_{\mathbf{d}}$, already considered
$\mathcal{T}$ and corresponding optimal DNN designs
$\mathcal{X}^*=\{\mathbf{x}^*(t), \forall t\in\mathcal{T}\}$, small
$\delta>0$ for checking latency constraint satisfaction,
and maximum iteration $Max\_Iterate$\\
    \textbf{Output: }
	Optimal DNN design $\mathbf{x}^*$\\
    \textbf{Initialize:} Set $t_{\min}=0$ and $t_{\max}=1$;\\
    \For{$i$= 1 to $Max\_Iterate$}{
        $t=\frac{t_{\min}+t_{\max}}{2}$;\\
        \If{$t\not\in\mathcal{T}$}{
        Solve \eqref{eqn:obj_new} and obtain $\mathbf{x}^*(t)$;\\
        $\mathcal{T}\leftarrow\mathcal{T}\cup\{t\}$ and
        $\mathcal{X}^*\leftarrow\mathcal{X}^*\cup\{\mathbf{x}^*(t)\}$
        }
        Measure latency $latency(\mathbf{x}^*(t^*);\mathbf{d})$;\\
        \uIf{$latency(\mathbf{x}^*(t^*);\mathbf{d})\geq \overline{L}_{\mathbf{d}} + \delta$}{
            $t_{\min}=t$;
        }
        \uElseIf{$latency(\mathbf{x}^*(t^*);\mathbf{d})\leq \overline{L}_{\mathbf{d}} - \delta$}{
            $t_{\max}=t$;
        }
        \uElse{Break;}
    }

	\Return $\mathbf{x}^*(t)$;
	\label{algo:new_algorithm}
\end{algorithm}

With performance monotonicity, the relative ranking
of different DNNs' latency performances is preserved between
the two devices.
For example,
 as shown in Fig.~4 in \cite{DNN_NAS_ChamNet_Prediction_CVPR_2019_dai2019chamnet},
for different convolution operators,
latency performance monotonicity is observed between Samsung S8
with Snapdragon 835 mobile CPU and
Hexagon v62 DSP with 800 MHz frequency, although the absolute
performances are very different. We also show in Fig.~\ref{fig:performance_monotonicity} the performance monotonicity
of a set of 40 DNN models with different architectures on Google
Pixel 1 and Pixel 2. These two devices have major differences
in terms of several specifications, such as
operating systems (Android 7.1 vs. Android 8.0),
chipset (Qualcomm MSM8996 Snapdragon 821 with 14 nm
vs. Qualcomm MSM8998 Snapdragon 835 with 10 nm),
CPU (Quad-core 2x2.15 GHz Kryo \& 2x1.6 GHz Kryo
vs. Octa-core 4x2.35 GHz Kryo \& 4x1.9 GHz Kryo)
and GPU (Adreno 530 vs Adreno 540), which can affect
the latencies.
As a result,
the absolute latency values on these
two devices are very different and not following a simple scaling relation.
Nonetheless, on these two devices, many of the DNNs
 preserve performance monotonicity very well. Moreover, we see
that the latency predictor built on Google Pixel 1 is quite accurate
compared to the true value.
This demonstrates
that the latency predictor on Google Pixel 1 can also be reused
for Pixel 2, although the authors build another latency predictor for Pixel~2
in their released files \cite{DNN_NAS_ProxylessNAS_LatencyPredictor_SongHan_ICLR_2019_cai2018proxylessnas}.

As a result, the latency constraint $latency(\mathbf{x};\mathbf{d})\leq \overline{L}_{\mathbf{d}}$ can be transformed into
$latency(\mathbf{x};\mathbf{d}_0)\leq \overline{L}'_{\mathbf{d}}$.
That is, there exists another latency constraint $\overline{L}'_{\mathbf{d}}$
such that if the latency of a DNN design $\mathbf{x}$
on the proxy device $\mathbf{d}_0$ satisfies $latency(\mathbf{x};\mathbf{d}_0)\leq \overline{L}'_{\mathbf{d}}$, then the latency of the same DNN design
$\mathbf{x}$ on our target device $\mathbf{d}$ will meet is actual
latency constraint, i.e., $latency(\mathbf{x};\mathbf{d})\leq \overline{L}_{\mathbf{d}}$.

Consequently, we convert the original latency constraint
$latency(\mathbf{x};\mathbf{d})\leq \overline{L}_{\mathbf{d}}$
into an equivalent latency constraint expressed on the proxy device
$latency(\mathbf{x};\mathbf{d}_0)\leq \overline{L}'_{\mathbf{d}}$,
 which we
 can reuse the proxy device's latency predictor to approximate
(i.e.,  $\overline{latency}_{\mathbf{d}_0}(\mathbf{x})\leq \overline{L}'_{\mathbf{d}}$).
Therefore, based on proxy device's predictor, the DNN design problem for our new target device
can be re-written as
\begin{eqnarray}\label{eqn:objective_new}
  \min_{\mathbf{x}\in\mathcal{X}} -Acc_{\Theta_A}(\mathbf{x}),\;\;\;\;\;
  s.t., \;\;\overline{latency}_{\mathbf{d}_0}(\mathbf{x})\leq \overline{L}'_{\mathbf{d}}.
\end{eqnarray}


Nonetheless, without knowing $\overline{L}'_{\mathbf{d}}$
a priori, we cannot directly solve the constrained optimization problem \eqref{eqn:objective_new}.
Thus, we reformulate the problem \eqref{eqn:objective_new} as
\begin{equation}\label{eqn:obj_new}
\min_{\mathbf{x}\in\mathcal{X}}-(1-t)\cdot Acc_{\Theta_A}(\mathbf{x})
+t\cdot \overline{latency}_{\mathbf{d}_0}(\mathbf{x}),
\end{equation}
where $t\in[0,1]$ plays an equivalent role as $\lambda_2$ in the original relaxed problem
in \eqref{eqn:obj_transformed}. With a larger value of $t$, the resulting latency
will be smaller (predicted for the proxy device), and vice versa.
Importantly, because of performance monotonicity, a larger $t$ will also result
in a smaller latency on the new target device.
Given each value of $t$, the problem \eqref{eqn:obj_new}
can be quickly solved (e.g., using search-based algorithms), because the objective
function can be efficiently evaluated based on accuracy/latency predictors built
on the proxy device. For each $t$, there exists a corresponding optimal $\mathbf{x}^*(t)$.

Now, the problem reduces to finding an optimal $t^*$ such that
the actual latency constraint $latency(\mathbf{x};\mathbf{d})\approx \overline{L}_{\mathbf{d}}$ is satisfied\footnote{If the latency constraint
is very loose (i.e., $\overline{L}_{\mathbf{d}}$ is sufficiently large),
then the actual latency $latency(\mathbf{x};\mathbf{d})$ will always
be smaller than $\overline{L}_{\mathbf{d}}$. In this case, we have
$t^*\to0$.} and the accuracy is also maximized (i.e., minimizing $-Acc_{\Theta_A}(\mathbf{x})$). Then, given $t^*$, we can obtain
$\mathbf{x}^*(t^*)$. Specifically, for each $t$, we measure the actual latency
$latency(\mathbf{x}^*(t^*);\mathbf{d})$ and check if it just meets the actual latency constraint
$\overline{L}_{\mathbf{d}}$.
Since $t$ is a scalar, we can efficiently search for the optimal $t^*$
using bi-section methods. For example, even with a granularity of 0.001 (i.e., 1001 possible
values of $t\in[0,1]$), we only need at most $10=\lceil\log_2(1001)\rceil$ searches and latency measurements on the target device. This can reduce the significant
cost of building a latency predictor for the target device.
The algorithm is described in Algorithm~\ref{algo:new_algorithm}.

\subsection{Remarks}

We offer the following remarks on our first approach.

\textbf{Proxy latency with monotonicity.}
Essentially, the proxy device's latency predictor
$\overline{latency}_{\mathbf{d}_0}(\mathbf{x})$ serves
as a proxy latency for the actual target device. Nonetheless,
a key novelty and difference from the FLOP-based proxy latency function  is that
$\overline{latency}_{\mathbf{d}_0}(\mathbf{x})$ can preserve performance
monotonicity for a large group of devices (i.e.,
a larger $\overline{latency}_{\mathbf{d}_0}(\mathbf{x})$
also means a large actual latency on the target device), whereas
FLOP-based proxy latency does not have this desired property and a higher
FLOP can commonly have a smaller latency on a target device.

\textbf{When performance monotonicity
does not hold.}  The core idea of our first approach is to leverage
the performance monotonicity of DNNs on different devices.
But, this may not hold for all devices: a DNN model with the lowest latency on
one device may not always have the best latency performance on another device \cite{DNN_Systems_ShuffleNetV2_Guideline_ArchitectureDesign_10.1007/978-3-030-01264-9_8}.
The violation of performance monotonicity can be found when the actual latency of a new DNN design becomes significantly higher while it is expected to be lower.
If the performance monotonicity does not hold between
the proxy device and the new target device, then we will train a new performance predictor
for the new target device and treat it as a new proxy device (for possible future reuse); when another device
arrives, we will match it with the best suitable proxy devices based on their similarities,
and if performance monotonicity does not hold between the new target device and any
of the existing proxy devices, we will train a new performance predictor for this new device.

Note that performance monotonicity is not required to strictly
hold
for all DNNs, as long as it \emph{approximately} holds for optimal DNN designs $\mathbf{x}^*(t)$ for a sufficiently large set of $t$. The reason is
that the DNN design problem is non-convex and we only expect to find
a reasonably good DNN design, rather than the truly global optimal design.
We expect performance monotonicity at least among a group of  devices that
are \emph{not} significantly different from each other (e.g., see Fig.~\ref{fig:performance_monotonicity}
for latencies on Google Pixel 1 and Pixel 2, which have different
operating systems, chipsets, CPUs and GPUs).

In any case, our approach will not be slower than the existing predictor-aided
DNN optimization that requires performance predictors for each different device \cite{DNN_NAS_ChamNet_Prediction_CVPR_2019_dai2019chamnet},
since our approach can always roll back to the existing approaches
by treating each target device as a new proxy device.

\textbf{Energy constraint.}  If we also want to factor energy into the objective function, we need to
consider a new objective function parameterized by $\mathbf{t}=(t_1,t_2)$ where $t_1\geq0$, $t_2\geq0$, and $t_1+t_2\leq1$:
\begin{equation}\label{eqn:obj_new_energy}
\min_{\mathbf{x}\in\mathcal{X}}-(1-t_1-t_2)\cdot Acc_{\Theta_A}(\mathbf{x})
+t_1\cdot \overline{latency}_{\mathbf{d}_0}(\mathbf{x}) + t_2\cdot \overline{energy}_{\mathbf{d}_0}(\mathbf{x}),
\end{equation}
 where  $\overline{energy}_{\mathbf{d}_0}(\mathbf{x})$
 is the proxy device's energy predictor.
Accordingly, we need to extend Algorithm~\ref{algo:new_algorithm} to consider a search process over $t_1$ and $t_2$. While this is more complicated than bi-section on a scalar value,
there exist efficient search methods over a multi-dimension space \cite{Bisection_Multivariate_arXiv_2017_galvan2017multivariate}.
Regardless, searching over a low-dimensional parameter space $(t_1,t_2)$
is much easier than searching over the DNN design space (e.g., architecture space).

\section{Approach 2: Learning to Optimize}

\subsection{Overview}

While our first approach aims at avoiding training performance
predictors for each individual device, we still
need to take a small number of actual latency/energy measurements
on each target device, because the proxy device's performance predictor
can only provide a relative/ordered performance instead of the absolute performance.
To  scale up the optimization of DNNs for edge inference and generate
an optimal DNN design instantly for each target device, we now present
our second approach.

Our key idea is \emph{learning to optimize}: instead of performing
DNN design optimization repeatedly (once for an individual device),
we first learn a DNN \emph{optimizer} from DNN optimization
on sample devices,
and then apply the learnt DNN optimizer to new unseen devices and directly obtain
the optimal DNN design.

\begin{figure*}[!t]
    \centering
	\includegraphics[trim=0cm 4.2cm 0cm 0.0cm, page=2,clip,width=1\textwidth]{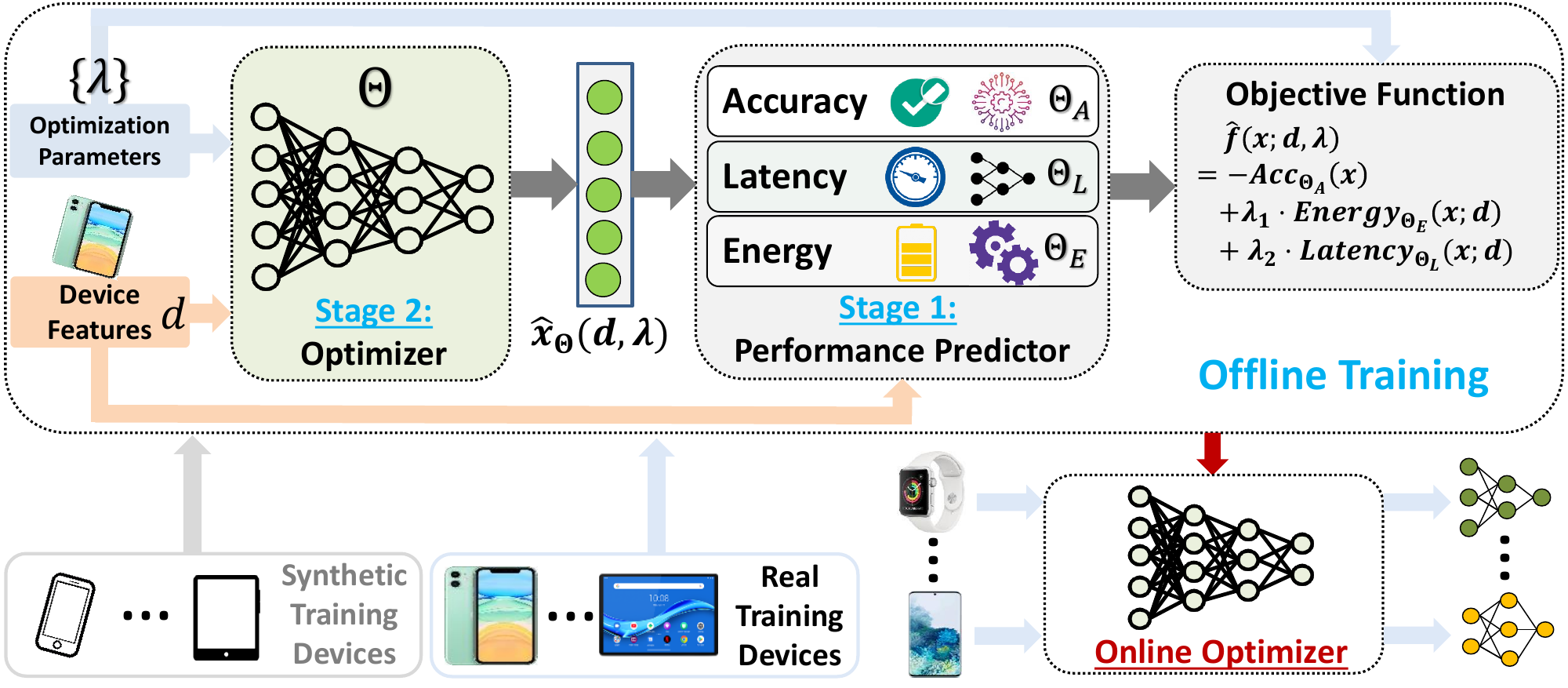}
	\caption{Overview of ``learning to optimize'' to scale up DNN optimization for edge inference.
Once the optimizer is trained, the optimal DNN design for a new device is done
almost instantly (i.e., only one inference time).}
	\label{fig:overview}
\end{figure*}

More specifically, we take a departure from the existing practice by:
(1) leveraging new performance predictors that can
estimate the resulting inference latency/energy performance
given a DNN-device pair;
and (2) using an automated optimizer which takes the device features
and optimization parameters as input, and then directly outputs the optimal
DNN design. This is illustrated in Fig.~\ref{fig:overview}.
Our latency/energy performance predictors take as explicit input both the
DNN features and device features, and hence they  can output the resulting performance for new unseen devices. Note that appropriate embedding of DNN and device features will be very helpful to facilitate training the performance predictors and DNN optimizer.

Our automated optimizer utilizes a neural network
to approximate the optimal DNN design function, and is intended to cut the search time
that would otherwise be incurred for each device.
The initial
overhead of training our performance predictors and optimizer
is admittedly higher than the current practice
of only training device-specific predictors, but the overall
overhead is expected to be significantly lower, considering the extreme diversity
of edge devices.

\subsection{Training Performance Predictors and Optimizer}

Our proposed design builds on top of  two-stage  training as described  below.

\textbf{Stage 1: Training performance predictors}.
The accuracy predictor is the same as the
one used in our first approach, since it is measured
on a reference dataset without dependence on devices.
 On the other hand, the latency/energy
predictor neural network will use both device feature $\mathbf{d}$ and DNN design representation $\mathbf{x}$
as input, and output the respective performance. They are each trained
by running DNNs with sampled designs on training devices and using mean squared error (i.e.,
the error between the predicted performance and the true measured value)
as the loss function.
The key difference between our design and \cite{DNN_NAS_APQ_JointSearch_ArchitecturePruningQuantization_SongHan_CVPR_2020_Wang_2020_CVPR,DNN_NAS_ChamNet_Prediction_CVPR_2019_dai2019chamnet}
is that our latency/energy performance predictors use device features
as part of the input and hence can apply to new unseen devices without training
new performance predictors.

We denote the set of training edge device features as $\mathcal{D}_T'$, where each
element $\mathbf{d}\in\mathcal{D}_T'$ corresponds to the feature of one available training device.
To generate  training samples, we can randomly sample some DNN designs
(e.g., randomly select some architectures)
plus existing DNN designs if available,
and then measure their corresponding performances on training devices as the labels.
We denote
the trained accuracy/energy/latency predictor neural network
by $Acc_{\Theta_A}(\mathbf{x})$, $Energy_{\Theta_E}(\mathbf{x};\mathbf{d})$,
and $Latency_{\Theta_L}(\mathbf{x};\mathbf{d})$, respectively,
where $\Theta_A$, $\Theta_E$, and $\Theta_L$ are learnt parameters
for the three respective  networks.
Thus, the predicted objective function $\hat{f}(\mathbf{x};\mathbf{d},\lambda)$
can be expressed as
\begin{equation}\label{eqn:obj_predicted}
\begin{split}
\hat{f}(\mathbf{x};\mathbf{d},\lambda)=-Acc_{\Theta_A}(\mathbf{x})+\lambda_1\cdot Energy_{\Theta_E}(\mathbf{x};\mathbf{d})+\lambda_2\cdot Latency_{\Theta_L}(\mathbf{x};\mathbf{d}).
\end{split}
\end{equation}
The accuracy/energy/latency predictor neural networks are called \emph{performance
networks}, to be distinguished from the {optimizer network} we introduce below.

Since collecting energy/latency performances on
real training devices is time-consuming,
we can use iterative training to achieve better sample efficiency.
Specifically, we can first choose a small training
set of DNN designs at the beginning, and then iteratively include an exploration set of new DNN designs $\mathcal{X}_{explore}$  to update the performance networks.
This is described in Algorithm~\ref{algo:iterative_algorithm}.
The crux  is how to choose the exploration set $\mathcal{X}_{explore}$.
Some prior studies have considered Bayesian optimization to
balance exploration vs. exploitation
\cite{DNN_NAS_Predictor_BayesianOptimization_Oxford_2020_ru2020neural,DNN_NAS_Predictor_TongZhang_2019_shi2019multi},
and we leave the choice of $\mathcal{X}_{explore}$ in each iteration as our future work.

\textbf{Stage 2: Training the automated optimizer.} Given an edge device
represented by feature $\mathbf{d}$
and optimization parameter $\lambda$, the representation
of the corresponding optimal DNN design
can be expressed as a function $\mathbf{x}^*(\mathbf{d},\lambda)$.
The current
practice of DNN optimization is to repeatedly run an optimizer
(e.g., search-based algorithm), once for a single device,
to minimize the predicted objective
function \cite{DNN_NAS_APQ_JointSearch_ArchitecturePruningQuantization_SongHan_CVPR_2020_Wang_2020_CVPR,DNN_NAS_ChamNet_Prediction_CVPR_2019_dai2019chamnet}.
Nonetheless, obtaining $\mathbf{x}^*(\mathbf{d},\lambda)$ is non-trivial for each device and not scalable
to extremely diverse edge devices.
Thus,
we address the scalability issue by leveraging the strong prediction power of another fully-connected neural network
parameterized by $\Theta$ to approximate the optimal DNN design function  $\mathbf{x}^*(\mathbf{d},\lambda)$. We call this neural network \emph{optimizer network},
whose output is denoted by $\hat{\mathbf{x}}_{\Theta}(\mathbf{d},\lambda)$ where $\Theta$ is the network
parameter that needs to be learnt.
Once $\Theta$ is learnt, when a new device arrives, we can directly predict the corresponding optimal
DNN design choice $\hat{\mathbf{x}}_{\Theta}(\mathbf{d},\lambda)$.

For training purposes, in addition to features of real available training devices $\mathcal{D}_T'$, we can also generate a set of additional \emph{synthetic} device features $\mathcal{D}_S$
to augment the training samples.
We denote the combined set of  devices for training as $\mathcal{D}_T=\mathcal{D}_T'\cup\mathcal{D}_S$, and the training
set of optimization parameters as $\Lambda_T$ which is chosen according to practical needs (e.g., latency
may be more important than energy or vice versa).
Next, we discuss two different methods to train the optimizer network.

\begin{algorithm}[!t]
	\caption{Training Performance and Optimizer Networks}
	\textbf{Input: }
	Real training devices $\mathcal{D}_T'$, synthetic training
devices $\mathcal{D}_S$, training set of optimization parameters $\Lambda_T$, trained DNN models and their corresponding design space $\mathcal{X}$,  initial exploration set
of $\mathcal{X}_{explore}$, initial training sets of sampled DNN designs $\mathcal{X}_T\subset\mathcal{X}$ and the corresponding
accuracy/energy/latency labels measured on real training devices, and maximum iteration rounds $Max\_Iterate$\\
    \textbf{Output: }
	Performance network parameters $\Theta_A,\Theta_E,\Theta_L$, and optimizer
network parameter $\Theta$\\
    \textbf{Initialize:} Randomize $\Theta_A,\Theta_E,\Theta_L$, and $\Theta$;\\
    \For{$i$ = 1 to $Max\_Iterate$}{
        \For{$\mathbf{x}\in\mathcal{X}_{explore}
        \subset\mathcal{X}$ and $\mathbf{d}\in\mathcal{D}'_T$}{
                $\mathcal{X}_T\leftarrow\mathcal{X}_T\cup\{\mathbf{x}\}$;\\
                Measure $accuracy(\mathbf{x})$ for a new accuracy label;\\	
                Measure $energy(\mathbf{x};\mathbf{d})$ and $latency(\mathbf{x};\mathbf{d})$
                for new energy and latency labels, respectively;\\
                Update $\Theta_A,\Theta_E$, and $\Theta_L$ by training performance
                networks as described
                in \textbf{Stage 1};
        }
        Choose a new $\mathcal{X}_{explore}$;\\
    }
    \uIf{Training method 1 is used}{
       Fix $\Theta_A,\Theta_E,\Theta_L$, and obtain $\mathbf{x}^*(\mathbf{d},\lambda)=\arg\min_{\mathbf{x}}\hat{f}(\mathbf{x};\mathbf{d},\lambda)$, $\forall (\mathbf{d},\lambda)\in(\mathcal{D}_T,\Lambda_T)$;\\
        Update $\Theta$ by training the optimizer network using Method 1;
    }
    \uElse
    {
        Fix $\Theta_A,\Theta_E,\Theta_L$, and update $\Theta$ by training the optimizer network using Method 2;
    }

	\Return $\Theta_A,\Theta_E,\Theta_L$, and $\Theta$;
	\label{algo:iterative_algorithm}
\end{algorithm}

\hspace{0.5cm}\textbf{Training Method 1:} A straightforward method of training the optimizer
network is to use the optimal DNN design $\mathbf{x}^*(\mathbf{d},\lambda)$
as the ground-truth label for input sample $(\mathbf{d},\lambda)\in(\mathcal{D}_T,\Lambda_T)$. Specifically,
we can use the mean squared error loss
\begin{equation}\label{eqn:training_optimizer_1}
\begin{split}
&\min_{\Theta}\frac{1}{N}\sum_{(\mathbf{d},\lambda)\in(\mathcal{D}_T,\Lambda_T)}|\hat{\mathbf{x}}_{\Theta}(\mathbf{d},\lambda)-{\mathbf{x}}^*(\mathbf{d},\lambda)|^2
+\mu\|\Theta\|,
\end{split}
\end{equation}
where $N$ is the total number of training samples, $\mu\|\Theta\|$ is the regularizer to avoid over-fitting,
and the ground-truth optimal DNN design $\mathbf{x}^*(\mathbf{d},\lambda)$
is obtained by using an existing optimization algorithm (e.g., evolutionary search in \cite{DNN_NAS_APQ_JointSearch_ArchitecturePruningQuantization_SongHan_CVPR_2020_Wang_2020_CVPR,DNN_NAS_ChamNet_Prediction_CVPR_2019_dai2019chamnet}) based on the predicted
objective function.
Concretely, the optimal DNN design used as the ground truth is
$\mathbf{x}^*(\mathbf{d},\lambda)=\arg\min_{\mathbf{x}}\hat{f}(\mathbf{x};\mathbf{d},\lambda)$,
where $\hat{f}(\mathbf{x};\mathbf{d},\lambda)$ is the predicted objective
function with  parameters $\Theta_A$, $\Theta_E$, and $\Theta_L$ learnt
in Stage~1.

\hspace{0.5cm}\textbf{Training Method 2:} While Method 1 is intuitive,
generating many training samples by obtaining
the optimal DNN design
$\mathbf{x}^*(\mathbf{d},\lambda)$,
even based on the predicted objective function, can be slow \cite{DNN_NAS_APQ_JointSearch_ArchitecturePruningQuantization_SongHan_CVPR_2020_Wang_2020_CVPR,DNN_NAS_ChamNet_Prediction_CVPR_2019_dai2019chamnet}.
To reduce the cost of generating training samples,
we can directly minimize the predicted
objective function $\hat{f}(\mathbf{x};\mathbf{d},\lambda)=-Acc_{\Theta_A}(\mathbf{x})+\lambda_1\cdot Energy_{\Theta_E}(\mathbf{x};\mathbf{d})+\lambda_2\cdot Latency_{\Theta_L}(\mathbf{x};\mathbf{d})$ in an {unsupervised} manner,
without using the optimal DNN design choice $\mathbf{x}^*(\mathbf{d},\lambda)$ as the ground-truth {label}.
Specifically, given the input samples $(\mathbf{d},\lambda)\in(\mathcal{D},\Lambda)$
including both real and synthetic device features, we optimize the optimizer network parameter $\Theta$ to directly minimize the following loss:
\begin{equation}\label{eqn:training_optimizer}
\begin{split}
&\min_{\Theta}\frac{1}{N}\sum_{(\mathbf{d},\lambda)\in(\mathcal{D}_T,\Lambda_T)}\hat{f}(\hat{\mathbf{x}}_{\Theta}(\mathbf{d},\lambda);\mathbf{d},\lambda)
+\mu\|\Theta\|.
\end{split}
\end{equation}
The output of the optimizer network
directly minimizes the predicted objective function, and hence represents
the optimal DNN design.
  Thus, our training of the optimizer network in Method 2 is  guided
by the predicted objective function only and unsupervised.
When updating the optimizer network parameter $\Theta$,
the  parameters for performance predictors $\Theta_A$, $\Theta_E$, and $\Theta_L$ learnt
in Stage 1
are fixed without updating. In other words, by viewing the concatenation
of optimizer network and performance predictor networks  as a single neural network
(illustrated in Fig.~\ref{fig:overview}), we update the parameters ($\Theta$)
in the first few layers  while freezing
the parameters ($\Theta_A,\Theta_E,\Theta_L$) in the last few layers
to minimize the loss expressed in Eqn.~\eqref{eqn:training_optimizer}.

Finally, we can search for appropriate weight parameters $\lambda$ to obtain
the optimal DNN design subject to performance requirement. The key difference between
our second approach and the first one is that in the second approach, there is no
need to measure the performance for each candidate DNN design on the target device.
Note that in our first approach, for each target device, there are only
a few  candidate DNN designs due to the high efficiency bisection methods.

\subsection{Remarks}

In this section, we propose a new approach
to scaling up DNN optimization for edge inference
and present an example of training the optimizer.
The key point we would like to highlight in this work is that performing DNN optimization for each individual
device as considered in the existing research is not scalable
in view of extremely diverse edge devices.
We now offer the following remarks (mostly regarding our second
approach --- learning to optimize).

$\bullet$ \textbf{DNN update.} When a new training dataset is available and the DNN models need to be updated
for edge devices,
we only need to build a new accuracy predictor on (a subset of) the new dataset
and re-train the optimizer network. The average energy/latency predictors
remain unchanged, since they are not much affected by training datasets.
Thus, the time-consuming part of building energy/latency predictors
in our proposed approach is a one-time effort and can be re-used for future tasks.

$\bullet$ \textbf{Generating optimal DNN design.} Once the optimizer network is trained, we can directly generate the optimal
DNN design represented by $\hat{\mathbf{x}}_{\Theta}(\mathbf{d},\lambda)$ given
a newly arrived edge device $\mathbf{d}$ and optimization parameter $\lambda$.
Then, the representation $\hat{\mathbf{x}}_{\Theta}(\mathbf{d},\lambda)$ is
mapped to the actual DNN design choice using the learnt decoder.  Even though the optimizer network may not always result in the optimal DNN
designs for all edge devices, it can at least help
us narrow down the DNN design to a much smaller space, over which fine tuning
the DNN design becomes much easier than over a large design space.

$\bullet$ \textbf{Empirical effectiveness.} Using  performance predictors to guide the optimizer
is relevant to \emph{optimization from samples} \cite{OptimizationSample_LimitationsOfOPS_STOC_2017_6447_10.1145/3055399.3055406,OptimizationSample_PowerOfOPS_NIPS2016_6447}.
While in theory optimization from samples may result in bad outcomes because the predictors
may output values with significant errors,
the existing NAS and compression approaches using performance
predictors \cite{DNN_NAS_Predictor_NAO_NIPS_2018_luo2018neural,DNN_NAS_Predictor_Survey_YuWang_Tsinghua_2020_ning2020surgery,DNN_NAS_OnceForAll_SongHan_MIT_2020_ICLR,DNN_NAS_APQ_JointSearch_ArchitecturePruningQuantization_SongHan_CVPR_2020_Wang_2020_CVPR,DNN_NAS_ChamNet_Prediction_CVPR_2019_dai2019chamnet}
have empirically shown that such optimization from samples work very well and
are able to significantly improve DNN designs in the context of DNN optimization. This is partly due to the
fact that the predicted objective function only serves as a guide and hence does
not need to achieve close to 100\% prediction accuracy.

$\bullet$ \textbf{Relationship to the existing
approaches.} Our proposed design advances the existing prediction-assisted DNN optimization
approaches \cite{DNN_NAS_APQ_JointSearch_ArchitecturePruningQuantization_SongHan_CVPR_2020_Wang_2020_CVPR,DNN_NAS_ChamNet_Prediction_CVPR_2019_dai2019chamnet}
by making the DNN optimization process scalable to numerous diverse edge devices.
If our approach is applied to only \emph{one} edge device, then it actually reduces
to the methods in \cite{DNN_NAS_APQ_JointSearch_ArchitecturePruningQuantization_SongHan_CVPR_2020_Wang_2020_CVPR,DNN_NAS_ChamNet_Prediction_CVPR_2019_dai2019chamnet}.
Specifically, since the device feature $\mathbf{d}$
is fixed given only one device,
we can remove it from our design illustrated in Fig.~\ref{fig:overview}.
As a result, our performance predictors are the same as those in \cite{DNN_NAS_APQ_JointSearch_ArchitecturePruningQuantization_SongHan_CVPR_2020_Wang_2020_CVPR,DNN_NAS_ChamNet_Prediction_CVPR_2019_dai2019chamnet}.
Additionally, our optimizer network can be eliminated, or reduced to a trivial network that has
a constant input neuron directly connected to the output layers without any hidden layers.
Thus, when there is only one edge device, our approach is essentially identical to those in \cite{DNN_NAS_APQ_JointSearch_ArchitecturePruningQuantization_SongHan_CVPR_2020_Wang_2020_CVPR,DNN_NAS_ChamNet_Prediction_CVPR_2019_dai2019chamnet}.
Therefore, even in the worst event that the optimizer network or performance predictor
network does not generalize well to some new unseen edge devices (due to, e.g.,
poor training and/or lack of edge device samples), we can always
optimize the DNN design for each individual device, one at a time,
and
roll back to state of the art \cite{DNN_NAS_APQ_JointSearch_ArchitecturePruningQuantization_SongHan_CVPR_2020_Wang_2020_CVPR,DNN_NAS_ChamNet_Prediction_CVPR_2019_dai2019chamnet}
 without additional penalties.

$\bullet$ \textbf{When scalability is not needed.}
It has been widely recognized that a single DNN model cannot
perform the best on many devices, and device-aware DNN optimization
is crucial \cite{DNN_NAS_APQ_JointSearch_ArchitecturePruningQuantization_SongHan_CVPR_2020_Wang_2020_CVPR,DNN_NAS_OnceForAll_SongHan_MIT_2020_ICLR,DNN_NAS_ChamNet_Prediction_CVPR_2019_dai2019chamnet,DNN_NAS_Hardware-awareTransformer_Lanugage_SongHan_ACL_2020_wang2020hat,DNN_Facebook_Inference_HPCA_2019}.
Thus, we focus on the \emph{scalability} of DNN optimization
for extremely diverse edge devices.
On the other hand, if there are only a few target devices (e.g., a vendor develops its own specialized DNN model for only a few
products), our second approach does not apply
while our first appraoch (i.e., re-using proxy device's performance predictors is more suitable).

$\bullet$ \textbf{GAN-based DNN design.}  There have been recent attempts to reduce the DNN design space by
training generative adversarial networks  \cite{DNN_NAS_GenerativeDesign_Quantization_arXiv_2020_kao2020generative}.
Nonetheless, they only produce DNN design candidates that are more likely to satisfy
the accuracy requirement, and do not perform energy or latency optimization for DNN designs. Thus, a scalable
performance evaluator is still needed to identify an optimal DNN design for diverse
edge devices. By contrast,
 our second approach is inspired by ``learning to optimize'' \cite{LearningOptimize_Gradient_Google_NIPS_2016_andrychowicz2016learning}:
 our optimizer network
takes almost no time (i.e., only one optimizer network inference)
to directly produce an \emph{optimal} DNN design, and can also produce multiple
optimal DNN designs by varying the optimization parameter $\lambda$ to achieve
different performance tradeoffs.

$\bullet$ \textbf{Ensemble.}  To mitigate potentially bad predictions produced
by our optimizer or performance networks, we can use an ensemble in our second approach.
For example, an ensemble of latency predictors can be used to smooth
the latency prediction, while an ensemble of the optimizer network can
be used to generate multiple optimal DNN designs, out of which we select
the best one based on (an ensemble of) performance predictors.

$\bullet$ \textbf{Learning to optimize.}  Our proposed optimizer network is relevant to the concept
of {learning to optimize} \cite{LearningOptimize_Gradient_Google_NIPS_2016_andrychowicz2016learning},
but employs a different loss function in Method 2 which does not utilize ground-truth optimal DNN
designs as labels.
The recent study \cite{LearningOptimize_PowerAllocation_CongShen_TCOM_2020_8922744}
considers related {unsupervised} learning to find  optimal power allocation
in an orthogonal problem context of multi-user wireless networks, but the performance is evaluated based on theoretical
formulas. By contrast, we leverage performance predictors to guide the training
of our optimizer network and use iterative training.

$\bullet$ \textbf{Public datasets for future research.}  Finally, the lack of access to many diverse edge devices is a practical challenge
that prohibits many researchers from studying or experimenting scalable DNN optimization for edge
inference. While  there are large datasets available
on $(architecture, accuracy)$ \cite{DNN_NAS_Benchmark_SurrogateBenchmarks_301_2020_siems2020nasbench301},
to our knowledge, {there do not exist similar publicly-available
datasets containing $(architecture, energy, latency, device)$ for a wide variety
of devices}.
If such datasets can be made available, they
will tremendously help researchers
build novel automated optimizers to scale up
the DNN optimization for heterogeneous edge devices, benefiting
every stakeholder in edge inference be it a gigantic player or a small start-up.

\bibliographystyle{plain}


\end{document}